\documentclass[OTHER]{cta-author}

\usepackage{hyperref}
\usepackage{indentfirst}
\usepackage{booktabs}
\usepackage{ulem}
\newcommand{\white}{\textcolor[rgb]{1.0,1.0,1.0}}
\newcommand{\tb}{\textcolor[rgb]{0.0,0.0,0.0}}

\begin{document}
%%%%%%%%%%%%%% title and the authors %%%%%%%%%%

\supertitle{\tb{Extended Abstract}}

\title{\vspace{-5pt}\tb{\Large{Diving with Penguins: Detecting Penguins and their Prey in Animal-borne Underwater Videos via Deep Learning}}}

%%%%%%%%%%%%%% RICH: Just suggested an author order here to be helpful, but not wedded to this at all if anyone wants to change it %%%%%%%%%%
\author{\tb{
	\au{Kejia Zhang$^{1,*}$},
	\au{Mingyu Yang$^{1}$},    
	\au{Stephen D. J. Lang$^{2}$},
    \au{Alistair M. McInnes$^{3}$},
    \au{Richard B. Sherley$^{2}$}, \\and
    \au{Tilo Burghardt$^{1}$}}
}

\address{\tb{
	\add{1}{Department of Computer Science, University of Bristol, Bristol, UK}
	\add{2}{Centre for Ecology and Conservation, University of Exeter, Penryn, Cornwall, UK}
    \add{3}{BirdLife South Africa, Cape Town, South Africa}
	\email{zhangkejia2019@163.com}}\vspace{-5pt}
}

%%%%%%%%%% end title and the authors %%%%%%%%%%

%%%%%%%%%%%%%% abstract %%%%%%%%%%

\begin{abstract}
	\tb{African penguins (\textit{Spheniscus demersus}) are an endangered species. Little is known regarding their underwater hunting strategies and associated predation success rates, yet this is essential for guiding conservation. Modern bio-logging technology has the potential to provide valuable insights, but manually analysing large amounts of data from animal-borne video recorders (AVRs) is time-consuming. In this paper, we publish an animal-borne underwater video dataset of penguins and introduce a ready-to-deploy deep learning system capable of robustly detecting penguins (mAP50@98.0\%) and also instances of fish (mAP50@73.3\%). We note that the detectors benefit explicitly from air-bubble learning to improve accuracy. Extending this detector towards a dual-stream behaviour recognition network, we also provide the first results for identifying predation behaviour in penguin underwater videos. Whilst results are promising, further work is required for useful applicability of predation behaviour detection in field scenarios. In summary, we provide a highly reliable underwater penguin detector, a fish detector, and a valuable first attempt towards an automated visual detection of complex behaviours in a marine predator. We publish the networks, the 'DivingWithPenguins' video dataset, annotations, splits, and weights for full reproducibility and immediate usability by practitioners.\vspace{-5pt}}
\end{abstract}

\maketitle

%%%%%%%%%% end abstract %%%%%%%%%%

%%%%%%%%%%%%%% main matter %%%%%%%%%%

\section{Introduction and Motivation}
\label{sec:intro}\vspace{-5pt}

\tb{\textbf{Scientific Importance.} 
Understanding how animals encounter and ingest food is important as foraging provides the energy for all metabolic processes. However, variation in foraging ability and how food is distributed across the environment mean that, particularly when food is scarce, some animals may expend too much energy for too little gain, with consequences for survival and breeding \cite{Wilson2018, Campbell2019}. \tb{While directly measuring prey capture is challenging — particularly in aquatic animals — the advent of animal biometrics\cite{kuehl2013} and animal-borne video recorders (AVRs)~\cite{mcinnes2019up} is now making such direct observations possible for the first time.}}

\begin{figure}[t]
    \centering\vspace{-3pt}
\includegraphics[width=240px,height=141px]{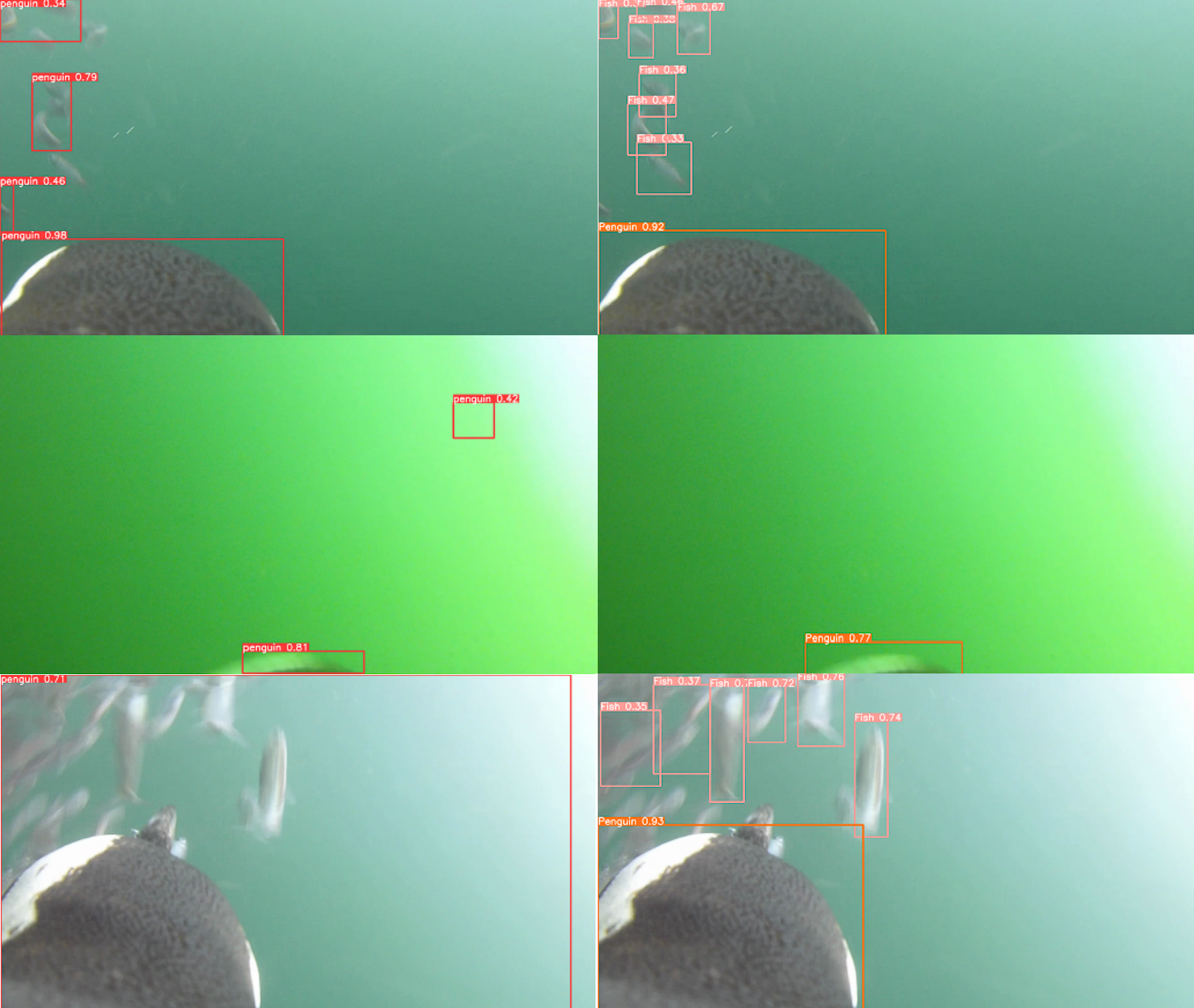}\vspace{-5pt}
    \caption{\tb{\textbf{Animal Detection in Penguin-borne Underwater Videos.} Three representative example frames of penguin-borne camera views from the DivingWithPenguins dataset. \textbf{\textit{(left)}} off-the-shelf YoloV5 penguin detection results, and \textbf{\textit{(right)}} results of our extended and fine-tuned multi-class model. Note qualitative improvements in localisation and classification performance. \vspace{-12pt}}}
    \label{fig:overview}
\end{figure}

\tb{\textbf{Conservational Motivation.} For African penguins (\textit{Spheniscus demersus}), which feed on shoaling fish off the coast of southern Africa, food scarcity has been implicated as a major factor in an ongoing and serious population decline. Between 1989 and 2019, the African penguin population declined by 65\% from c. 51,500 pairs to c. 17,700 pairs \cite{sherley2020conservation}, leading to the species being listed as globally Endangered in 2010 \cite{crawford2011collapse}. While various conservation strategies are being investigated for preserving the species, a lack of detailed information about prey availability and at-sea penguin foraging behaviour are hampering these efforts. To try to address some of these questions, researchers have begun to use AVRs~\cite{mcinnes2019up} to observe the foraging behaviour of African penguins (see~Fig.\ref{fig:overview}).} 

\tb{\textbf{Limitations of Manual Processing.} However, the vast amount of data produced by AVRs~ requires considerable time investment to manually annotate and classify. Further, when video data is annotated by multiple observers, there are likely to be subjective differences in the criteria individuals use to classify complex behaviours such as predation attempts. Machine learning methods~\cite{burg06,sherley2010spotting,kuehl2013,tuia2022} on the other hand operate fast and consistently and thus can help ecologists process visual data more efficiently and robustly~\cite{hughes,brust}}.

\tb{\textbf{Underwater AI Systems for Penguin Applications.} To date, several studies have proposed automated solutions to detect and classify the underwater behaviour of penguins. For example, time-series data from little penguins (\textit{Eudyptula minor}) wearing accelerometers has been matched with videos of the underwater behaviour of penguins in captivity to train a support vector machine model to {classify data from birds wearing accelerometers in the wild as swimming or prey handling with an accuracy of 84.95\% \cite{carroll2014supervised}. More recently, a deep learning neural network model has been developed to detect five different types of penguin dive behaviour at different depths and {one behaviour at the water's surface} in animal-borne video \cite{conway2021frame}. The study combined a CNN with an RNN to build a frame-by-frame spatial and temporal neural network for video classification with an 85.4\% accuracy \cite{conway2021frame}. And in 2022, researchers used AVRs to capture the foraging behaviour of chinstrap penguins (\textit{Pygoscelis antarcticus})}, and combined manual video annotations with depth data to train a Random Forest model to predict foraging activity from dive profiles~\cite{manco2022predicting}. Their model was more accurate in identifying dives with no prey (90.8\%) and prey swarms (84.0\%), but had limitations identifying dives with small numbers of individual prey (16.9\%).}

\textbf{{Paper Contribution.}} Here, we publish the annotated underwater DivingWithPenguins video and still image dataset and use computer vision and deep learning techniques to analyse the animal-borne African penguin video content. Methodologically, we first construct and evaluate a YOLOv5-based penguin detector that reliably locates the camera-carrying penguin and conspecifics (other penguins) in the captured underwater video. Secondly, we extend the system to form a dual-stream neural network that combines appearance and motion information towards the detection of challenging foraging behaviour of penguins in underwater video.

\vspace{20pt}
\section{The 'DivingWithPenguins' Dataset}
\label{sec:dataset}\vspace{-5pt}
\tb{\textbf{Dataset Overview.} We publish a video and still dataset that was captured using lightweight underwater cameras {temporarily mounted to the backs of breeding African penguins using Tesa Tape for a single foraging trip}~\cite{mcinnes2019up}. {The dataset contains a total of 63 videos from penguins equipped with cameras at the Stony Point breeding colony (34$^{\circ}$ 37' 14.21"S, 18$^{\circ}$ 89' 32.65" E), South Africa, in 2017}. Ethics protocols for animal handling and device application were fully observed when conducted by BirdLife South Africa. Based on the captured video dataset, we provide three task-specific data views and associated annotation metadata for training and evaluating animal, prey group, and behaviour recognition components. All splits, that is training and test data are made available with this paper for full transparency (see Section~\ref{sec:ack}).}

\vspace{-5pt}
\subsection{DATA VIEW A: Penguin Detection in Stills}\vspace{-5pt}
\tb{\textbf{Single Frame Object Detection Dataset.} 602 underwater images at a resolution of $640 \times 640$ pixels with a training portion of 418 images and a test portion of 184 images are designed to be utilised to train an underwater penguin detector. Following the requirements of YOLOv5 standards, this dataset view contains full bounding box annotations for penguins, fish, and bubbles in underwater images.}

\vspace{-5pt}
\subsection{DATA VIEW B: Content Classification}\vspace{-5pt}
\tb{\textbf{Whole Frame Classification Dataset.} For classifying frame content without object localisation a separate data and annotation format is provided: 797 underwater images at a resolution of $640 \times 640$ pixels with binary classification labels indicating the existence or absence of fish. In this data view, there are 370 images labelled as ``Have Fish'' and 427 images labelled as ``No Fish''. 558 images from the training portion and 239 images from the test portion. }

\vspace{-5pt}\subsection{DATA VIEW C: Behaviour Recognition}\vspace{-5pt}
\tb{\textbf{Short Behavioural Video Dataset.} In the full unprocessed  dataset of videos between 20min and 1h, some clips contain footage of penguins at the water surface and related unclear white splash clips. To limit video data to underwater behaviours original videos were cut into short videos which focus on underwater content and still contain all the predation events from the original video dataset. 188 predation events are annotated across 85 short videos and stored in a JSON file. Penguin feeding behaviours are fast actions that occur across about ten to twenty frames in the 30fps content. Hence, annotation is using milliseconds to pinpoint events accurately. Training and test instances are provided.}\vspace{-5pt}

\begin{figure}[b]
    \centering
                \includegraphics[width=115px,height=55px]{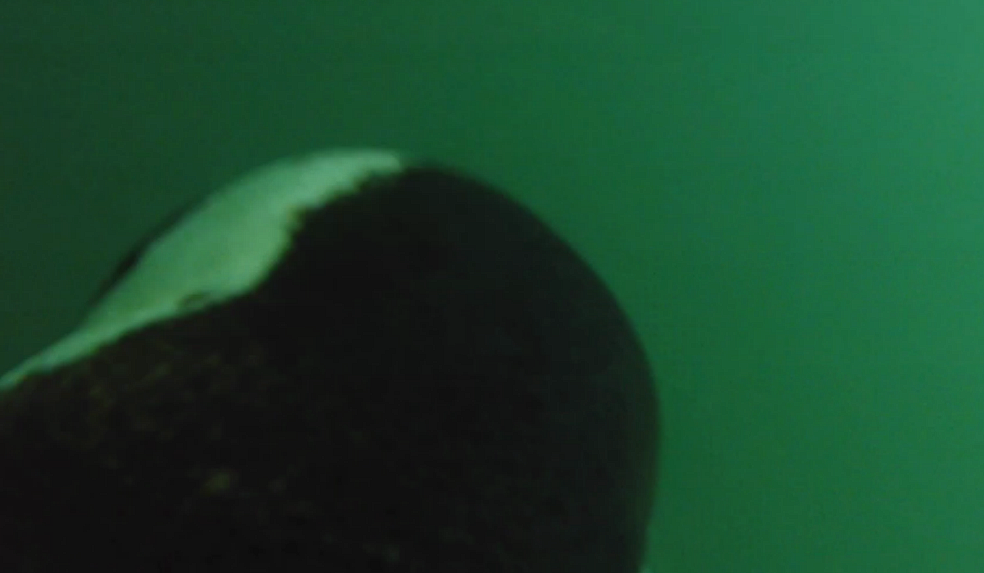}
                \includegraphics[width=115px,height=55px]{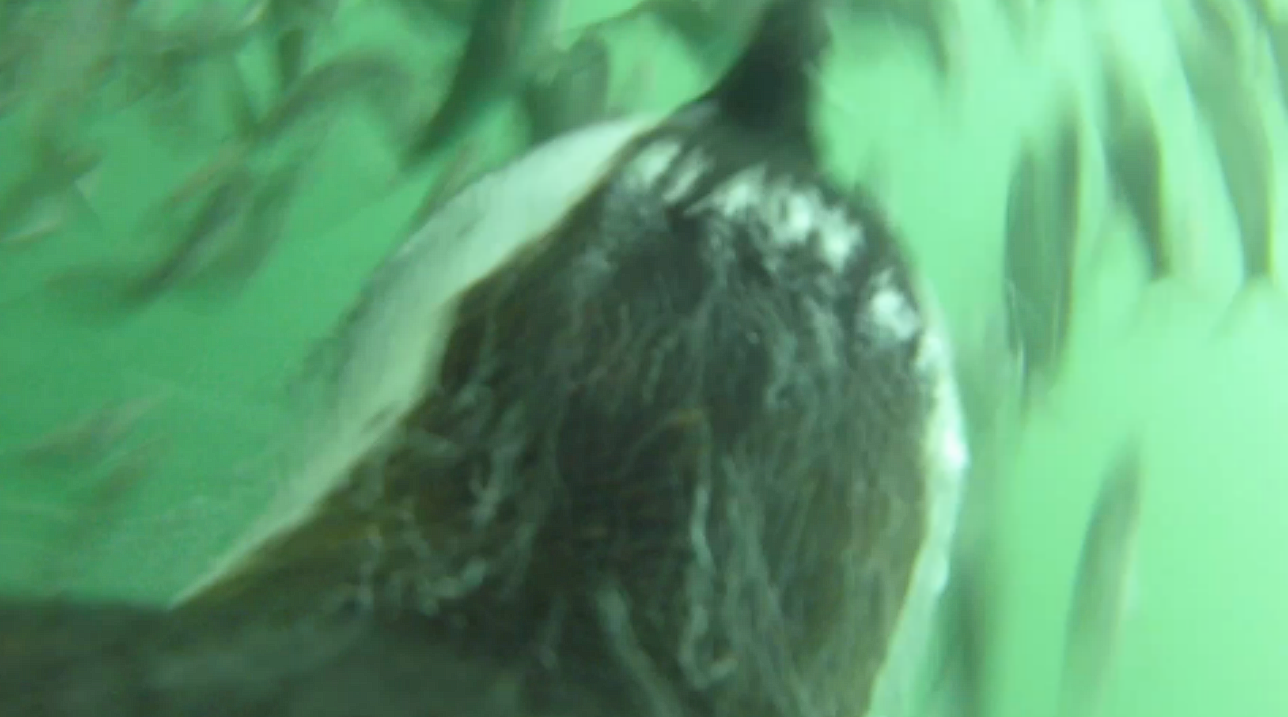}\vspace{3pt}
    \includegraphics[width=115px,height=55px]{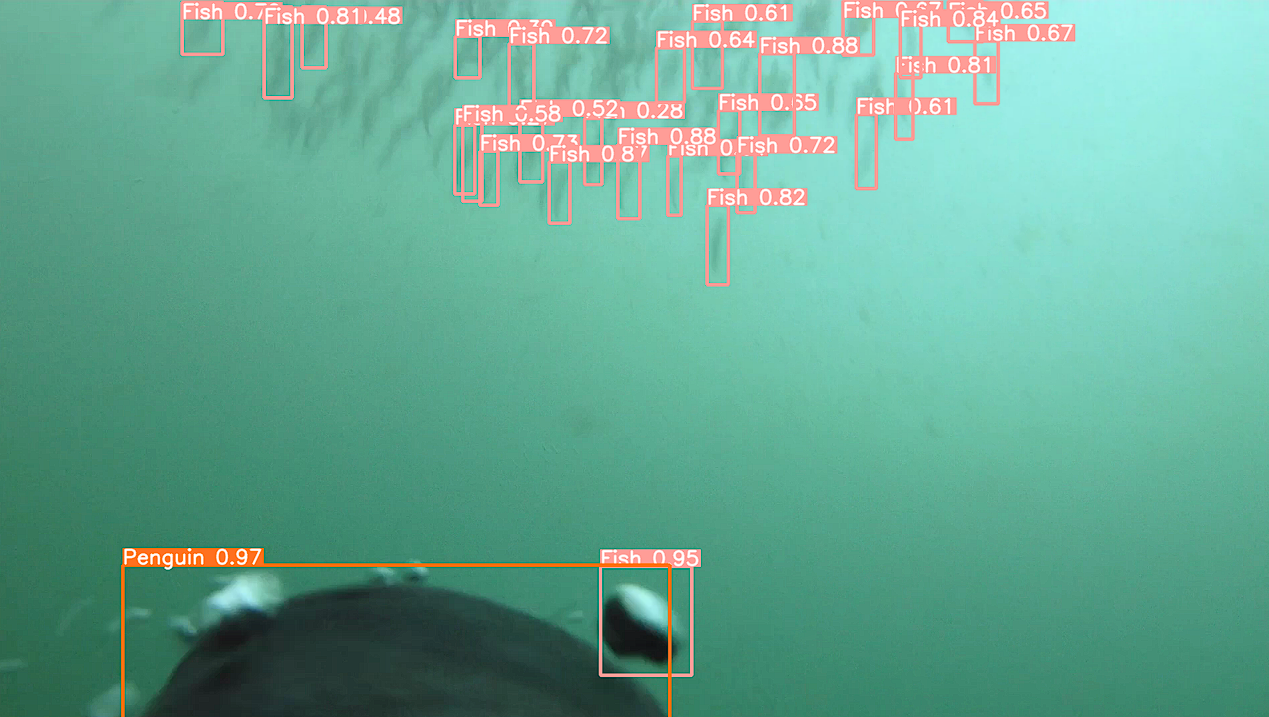}
        \includegraphics[width=115px,height=55px]{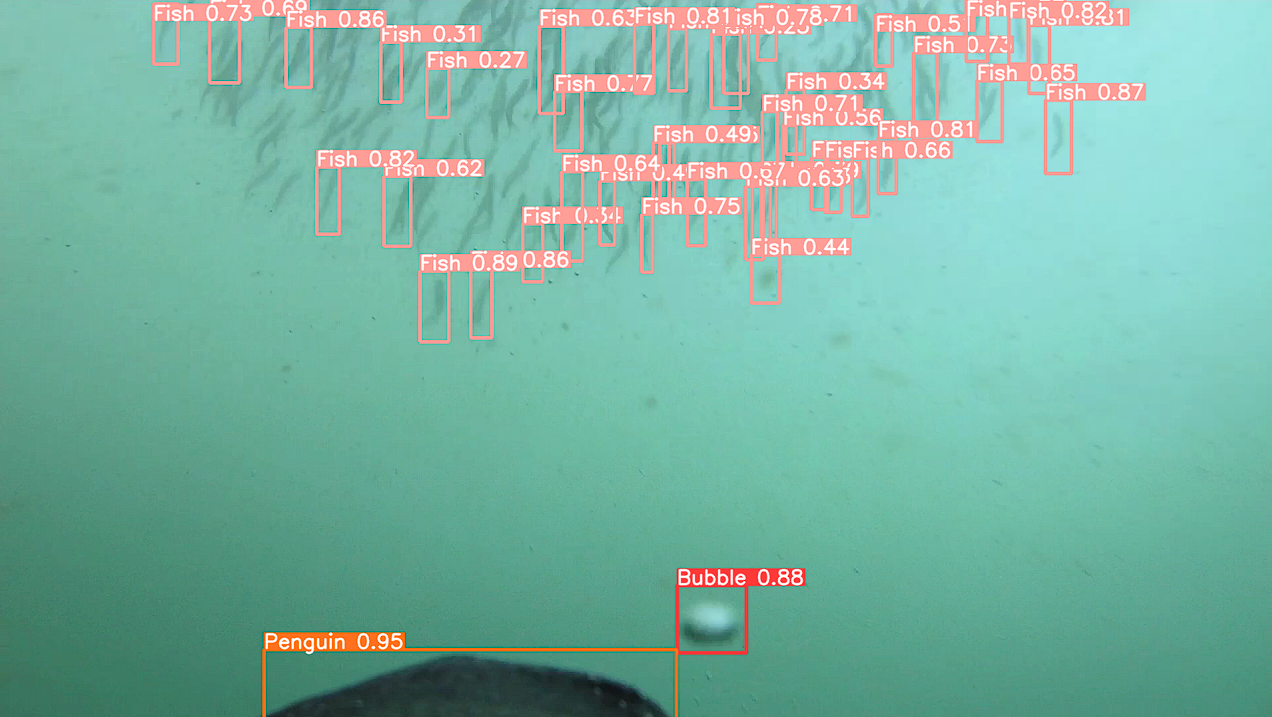}

    \caption{\tb{\textbf{Coping with Challenging Fish Visuals.} \textbf{\textit{(top left)}} clear frame with no fish; \textbf{\textit{(top right)}} dispersed 
 school of fish with non-distinct object boundaries of individual fish emphasised need for 'whole frame' recognition re fish presence; \textbf{\textit{(bottom left)}} poor fish object detection due dispersion of objects and distractor air bubble; \textbf{\textit{(bottom right)}} detection with explicit air bubble recognition.\vspace{-15pt}}}
    \label{fig:fish}
\end{figure}

\section{Method and Implementation}
\label{sec:method}\vspace{-5pt}

\subsection{YOLOv5 Penguin, Fish, and Bubble Detector}\vspace{-5pt}
\label{sec:penguin detection}
\tb{\textbf{Penguin Instance Localisation.} Our penguin detection model is implemented based on the YOLOv5 architecture developed by \textbf{Ultralytics}~\cite{ultralytics2021yolov5} and rooted in the classic real-time object detection algorithm YOLO~\cite{redmon2016you}. Performing a single forward pass of the image through the network produces two concurrent outputs: one for localisation and classification, respectively. In 2022, the first preliminary YOLOv5 baseline model~\cite{mingyu2023} was trained for the detection of penguins in underwater videos. The solution outperformed the MegaDetector v4 and v5 versions~\cite{mingyu2023} and forms the foundation of the detector presented here. Most critically, the baseline model (see~left images in Fig.\ref{fig:overview}) suffered from penguin misdetections  due to the presence of distractors such as fish or air bubbles. We address this by explicitly adding fish and air bubble categories to allow the detector explicitly to recognise and disambiguate these objects~(see~Fig.\ref{fig:fish}). Full model details are published in the code repository.}
\vspace{-8pt}
\subsection{Dual-Stream Predation Event Recognizer}
\begin{figure}
    \centering \vspace{-20pt}
    \includegraphics[width=240px,height=310px]{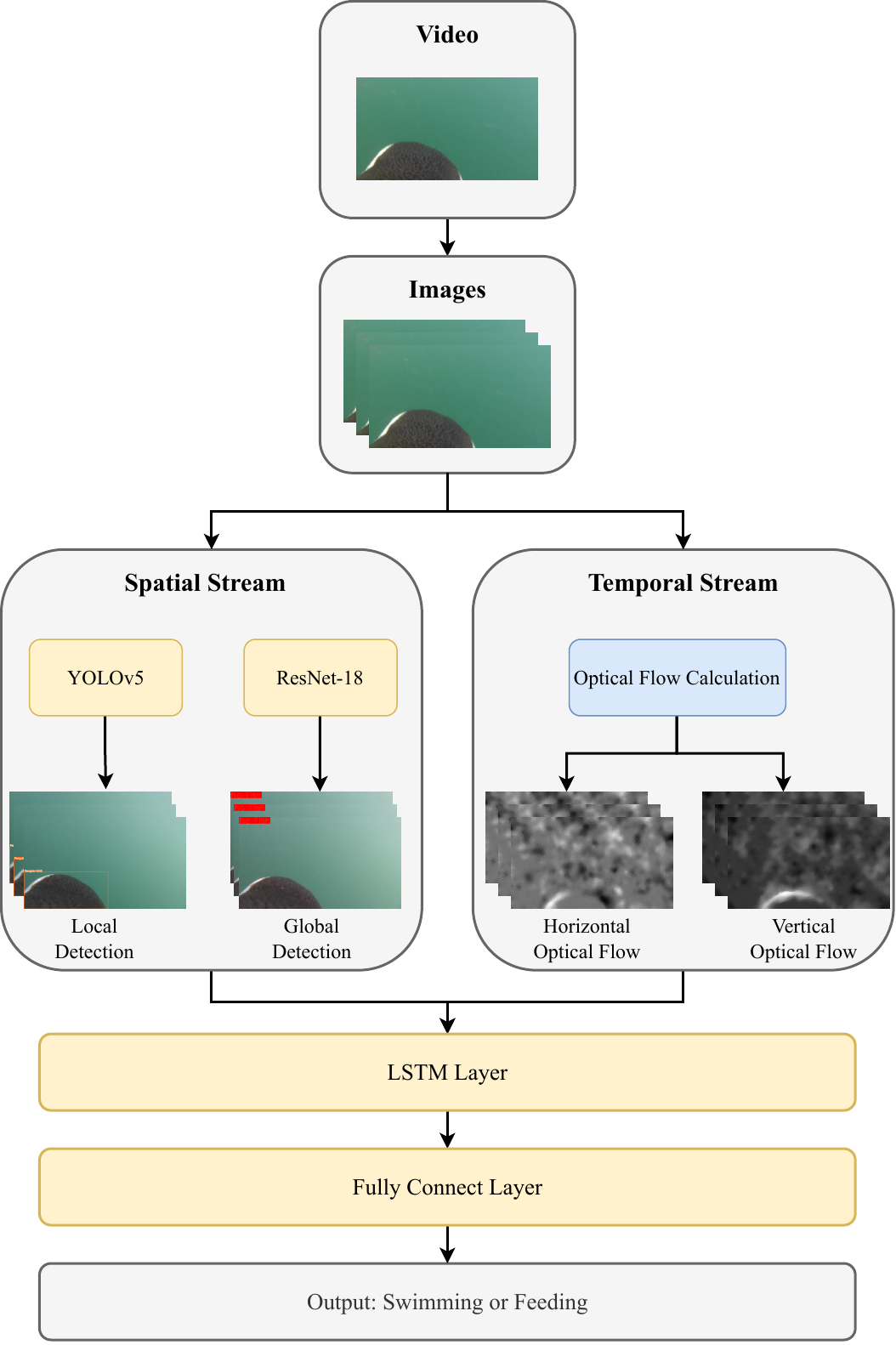}
    \caption{\tb{\textbf{Dual-Stream Behaviour Detection Network Design.} Short video snippets are processed frame-by-frame via spatial and temporal streams to capture RGB and optical flow information, respectively. Temporal integration across frames of the snippet is implemented via a LSTM layer. The final detection head outputs estimates whether the penguin carrying the camera is feeding or swimming.\vspace{-15pt}}}
    \label{fig:Dual}
\end{figure}

\vspace{-5pt}
\tb{\textbf{Finding Instances of Foraging Behaviour.} A dual-stream network structure is used as the backbone of the predation behaviour detector utilising appearance and motion information, respectively. Fig.~\ref{fig:Dual} shows a flow chart of the network design developed in this project. Note the spatial and temporal streams to capture RGB and optical flow information before temporal integration via a LSTM layer. A fully connected layer finally derives the semantic output indicating the behaviour of the penguin in the video. Full model details are published in this paper's code repository.}

\tb{\textbf{Spatial Stream for Local and Global Appearance Learning.}
The spatial network stream (see Fig.~\ref{fig:Dual}) consists of two fundamental subcomponents: 1) a YOLOv5 penguin detector as described in section \ref{sec:penguin detection} to provide object instance localisation features, and 2) a ResNet-18~\cite{he2016deep} frame classifier to provide whole-frame information about semantically important context (eg. indistinct presence of schools of fish). For the whole behavioural detection model, determining whether fish exist at all in a frame is an important precursor for finding penguin hunting behaviour. However, direct object detection has limitations regarding the identification of overlapping, large numbers of partially blurred fish. Adding a fish detection component as a global classifier without instance localisation addressed this issue. We used the TorchVision \textbf{torchvision.model} module of the PyTorch library with pre-trained weights from ImageNet~\cite{deng2009imagenet} as weight initialisation for training. In addition, since fish detection is a binary classification task, we changed the output dimension of the last fully connected layer in the ResNet-18~\cite{he2016deep} architecture to $1\times2$ representing the probability of fish presence or absence, respectively.}

\tb{\textbf{Temporal Stream for Motion Learning.}
We employ dense optical flow in vertical and horizontal directions as can be calculated from neighbouring frame pairs to estimate the normal flow motion at each pixel in the video. This calculation provides explicit dynamics information about the motion in the video. Optical flow is calculated via OpenCV's TV-L1 algorithm~\cite{perez2013tv} implementation using grayscale versions of any adjacent frame pair. The calculated motion vector for each pixel is finally decomposed into horizontal and vertical components and normalised (see Fig.~\ref{fig:Dual} for a visual).}

\begin{figure}[b]
    \centering\vspace{-18pt}
    \includegraphics[height=140px]{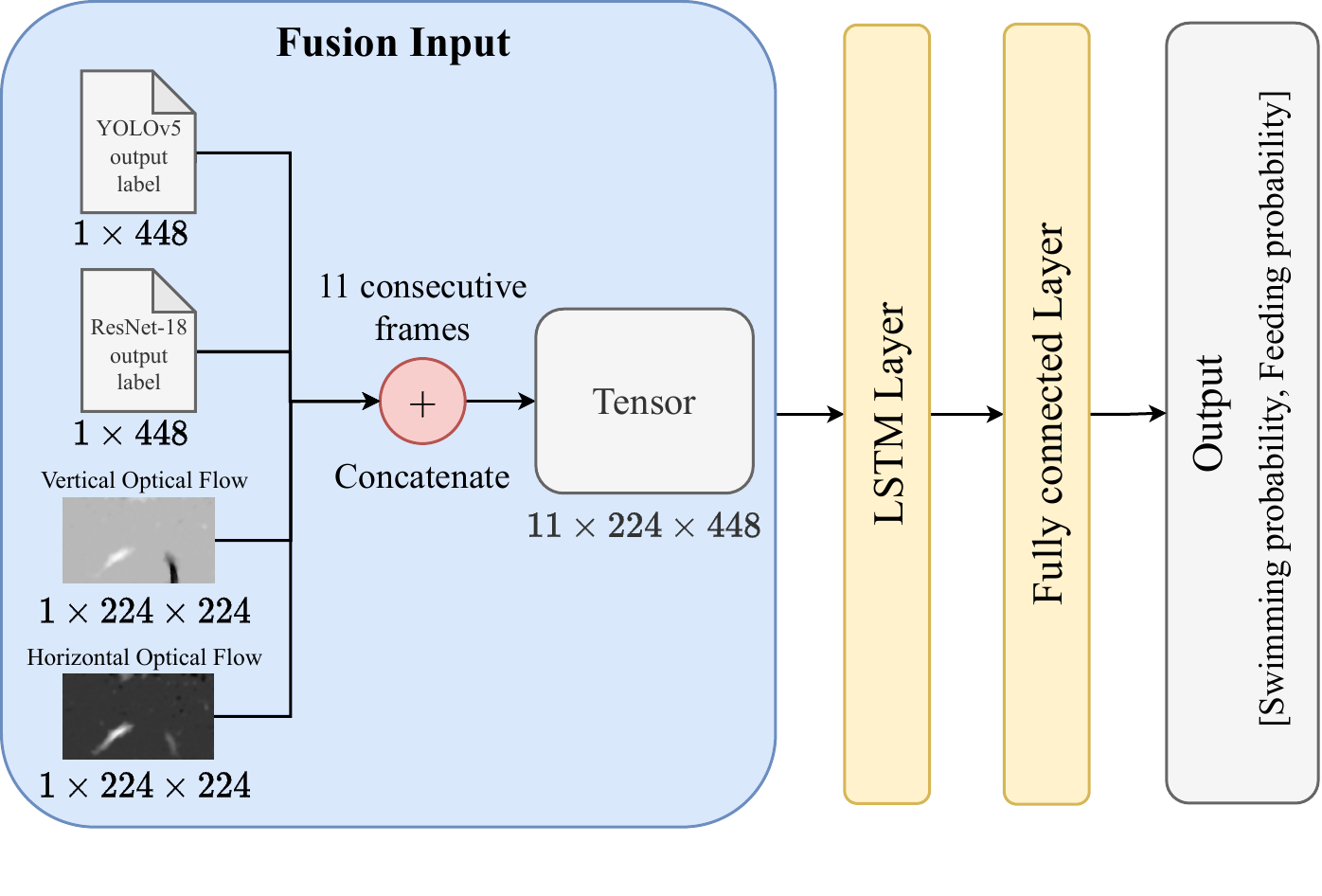}\vspace{-8pt}
    \caption{\tb{\textbf{LSTM Information Integration across Snippets.} Illustration of the integration process over 11 frames of result features across the two streams. Vectorized, concatenated and reshaped features are processed by an LSTM Layer to feed a classification head that makes judgements about predation behaviour based on information accumulated over an entire video snippet.\vspace{-12pt}}}
    \label{fig:LSTM input}
\end{figure}

\tb{\textbf{Multi-frame Integration via LSTM.} Behavioural animal detection in video~\cite{sakib,visapp23} requires incorporating and integration information distributed over many frames.  Our architecture utilises an LSTM layer~\cite{hochreiter1997long,andrew2017visual} to process groups of $n=11$ consecutive frames. Fig.~\ref{fig:LSTM input} illustrates the feature concatenation and frame integration process. For individual frames, the spatial detection results from the YOLOv5 and ResNet detector models as well as the horizontal and vertical optical flow images from the temporal stream are vectorized, concatenated, and converted into a tensor. This time series data feeds the LSTM model which integrates information and outputs an accumulated feature semantically interpreted by the classification head to form the network output.}

\begin{table*}[t]
\tb{\vspace{-20pt}
\begin{flushleft}
\begin{tabular}{@{}lllllllll@{}}
\toprule
        \textbf{Object Category}\white{----}& \multicolumn{4}{l}{\textbf{Off-the-Shelf YOLOv5 Penguin-only Detector~\cite{mingyu2023}\white{----}}}   & \multicolumn{4}{l}{\textbf{OUR YOLOv5 Multi-class Detector}}   \\ \midrule
        & Precision\white{---} & Recall\white{-----} & mAP50\white{-----}  & mAP50-95\white{-----------} & Precision\white{---} & Recall\white{-----}& mAP50 \white{----} & mAP50-95\white{---} \\
ALL     & 55.1\%    & 55.6\% & 60.6\% & 43.6\%   & 81.1\%    & 77.3\% & \textbf{81.1}\% & 52.5\%   \\
Penguin & 55.1\%    & 55.6\% & 60.6\% & 43.6\%   & 99.0\%      & 95.2\% & \textbf{98.0}\%   & 83.8\%   \\
Fish    & -         & -      & -      & -        & 70.2\%    & 70.0\%   & \textbf{73.3}\% & 32.2\%   \\
Bubble  & -         & -      & -      & -        & 74.2\%    & 66.8\% & \textbf{72.0}\%   & 41.8\%   \\ \bottomrule
\end{tabular}
\end{flushleft}
\caption{\tb{\textbf{Quantitative Object Detection Performance Statistics}. Comparison of \textbf{\textit{(left)}} off-the-shelf application of YOLOv5 for penguin detection as given in~\cite{mingyu2023} outperforming Megadetector v4 and v5~\cite{mingyu2023}  and \textbf{\textit{(right)}} our proposed YOLOv5 multi-class detector when evaluated on the DataViewA testing portion. Note the substantial performance improvements for penguin detection when explicitly learning the appearance of distractor classes such as fish or air bubbles.\vspace{-8pt}}}
\label{tab:yolo final_result}
}
\end{table*}

\tb{\textbf{Balanced Sampling.} We note that there is a significant imbalance in the number of rare predation and abundant non-predation swimming samples. Such imbalances often lead to model bias. To counteract this to some degree, non-predation events were randomly sampled at the same rate as predation samples occurred. Thus, in the training phase, all the predation event samples and the same number of non-predation samples are used.}
\vspace{-8pt}
\section{Experimental Results \& Evaluation}
\label{sec:evaluation}\vspace{-5pt}
\subsection{Enhanced Penguin, Fish, and Bubble Detection}\vspace{-5pt}
\tb{\textbf{Object Detection Performance.} All training in this paper is conducted on the Blue Crystal Phase 4 supercomputer at the University of Bristol. The main components used in the training are 2.4Ghz Intel E5-2680 v4 (Broadwell) CPUs and Nvidia P100 GPUs~\cite{bcp4}. After training YOLOv5 for 230 epochs via Stochastic Gradient Descent (SGD) on fish and penguin content in the training portion of DataViewA we find an mAP50 of 98\% for detecting penguins via off-the-shelf YOLOv5 outperforming the mAP50 of 60.6\% for penguin-only detection reported in previous Megadetector-beating work~\cite{mingyu2023}. Thus, explicit fish modelling can significantly improve penguin detection performance. Yet, an mAP50 of 63.1\% only can be achieved for 
fish. A major error source for fish detection of the network is the problem of mistakenly mixing up fish with 'undefined' and 'salient' background objects underwater. To address this, we explicitly added an air bubble category to the training in order to allow the model to better distinguish between the features of fish and the most dominant object-like background feature -- air bubbles. Heavy augmentation was used to maximise the visual sample space of bubbles. The full performance of this final model when trained with early stopping via the pocket algorithm is shown in Table \ref{tab:yolo final_result}. As can be seen, compared with the off-the-shelf baseline model, the accuracy has improved in detecting penguins and fish by applying the above methods. Note that separate fish detection applications could provide better insight into so far limited quantity estimates of prey -- see Sutton et al.~\cite{sutton22} for details.}

\begin{figure*}[t]
    \centering
    \includegraphics[width=518px, height=160px]{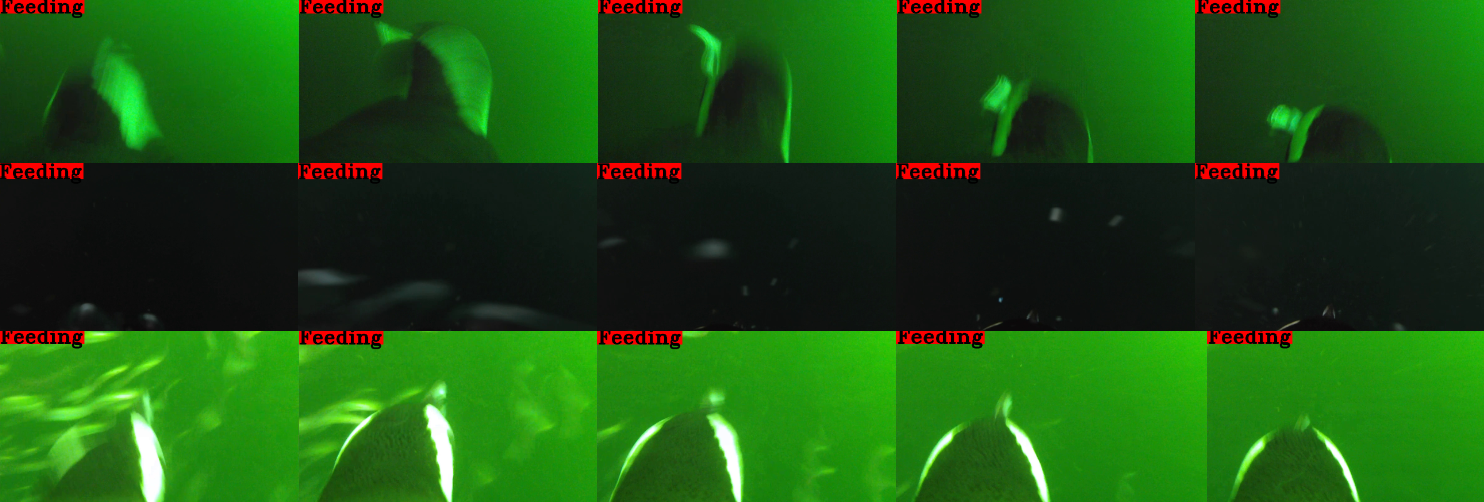}
    \caption{\tb{\textbf{Video Skims of Typical Highly Complex Behaviour Detection Content.} \textbf{\textit{(top)}} true positive example of a feeding sequence containing an actual feeding event; \textbf{\textit{(middle)}} misclassified rapid swimming sequence containing no foraging, but fast moving air bubbles around the beak area; \textbf{\textit{(bottom)}} misclassified sequence containing fish and air bubbles, however, no actual feeding takes place.\vspace{-5pt}}}
    \label{fig:wrong feeding}
\end{figure*}

\tb{\textbf{Confirming Practical Applicability.} To further evaluate performance qualitatively, we applied the penguin-only detector~\cite{mingyu2023} and our multi-class detector to the same underwater video sequences. Comparative detection results for three example frames are shown in Fig.~\ref{fig:overview}. Qualitatively, when observing performance across videos, we find that the number of false positive instances in the penguin category decreases drastically when using our multi-class penguin detector. Penguin detection is practically highly reliable when applied to unseen underwater videos confirming quantitative results and pointing towards effective usability in applications as is.}

\vspace{-5pt}
\subsection{Whole Frame Fish Classifier}\vspace{-5pt}
\tb{\textbf{Model Training.} We test ResNet-18 and ResNet-50 models with ImageNet pre-raining and perform a search over the basic space spanned by optimizer and learning rate. We test learning rates of 0.01, 0.001 and 0.0001 with SGD, SGD with different momenta, and ADAM. We found that a ResNet-18 model using the SGD optimiser with a learning rate of 0.0001 combined with 0.9 momentum performed the best. Table~\ref{tab:resnet baseline result} shows these best-performing parameter settings for a standard ResNet-18 fish classifier and its performance trained and tested on Data View B.}

\tb{\textbf{Model Optimisation and Performance.} We then apply further model optimisation methods, that is L2 regularisation, dropout, and geometric data augmentation (see code repository for details). However, we observe that only L2 normalization improves performance. Table~\ref{tab:resnet_summary} depicts the per-class performance of the off-the-shelf and optimised detector. We note that despite slightly improved overall per-class accuracy, improvements in fish detection are linked to higher sensitivity to misclassifications in NoFish frames. }

\vspace{-5pt}
\subsection{Spatio-Temporal Predation Behaviour Detector Trials}\vspace{-5pt}
\tb{\textbf{Feeding Recognition.} Accumulation of information in LSTM layers for behaviour recognition is centrally affected by two hyperparameters, the number of layers and the hidden layer size of the LSTM. The number of layers controls the complexity of the model, while the hidden layer size controls the size of the output dimension of the memory units in the LSTM. Tab.~\ref{tab:num_layer_hidden} shows test accuracy on the Data View C test portion for some key hyperparameter combinations. Testing code is provided in the repository for details how test snippets are accounted for. The results demonstrate that two-layer LSTMs show higher average accuracy regarding both feeding and swimming behaviour detection at fixed output dimension.}

\begin{table}[b]
\tb{\vspace{-15pt}
    \centering
    \begin{tabular}{@{}llllll@{}}
    \toprule
    \multicolumn{6}{c}{\textbf{\white{---}Best Training Config for ResNet-18 Fish Frame Detector\white{---}}}                                                    \\ \midrule
    Architecture              & \multicolumn{5}{c}{ResNet-18}                                                           \\
    Pre-trained               & \multicolumn{5}{c}{ImageNet}                                                                 \\
    Transfer learning         & \multicolumn{5}{c}{Fine-tune whole model}                                               \\
    Optimiser                 & \multicolumn{5}{c}{{SGD} + Momentum at 0.9}                                                      \\
    Loss                      & \multicolumn{5}{c}{Cross Entropy}                                                       \\
    Learning Rate             & \multicolumn{5}{c}{0.0001}\\ \hline               \end{tabular}\vspace{5pt}
\caption{\tb{\textbf{Training Parameterisation}. Listed are most important model, para-\\meter, and hyperparameter settings for off-the-shelf ResNet-18 models found \\as best-performing via parameter search and used for training the whole frame fish\\ classifier.}}
\label{tab:resnet baseline result}}
\end{table}

\tb{\textbf{Behaviour Recognition Limitations and Opportunities.} Despite acceptable quantitative results, we recognise that the behaviour detector is not yet ready for field application. By visualising the prediction results of the model, the true challenge of feeding recognition becomes apparent. Fig.~\ref{fig:wrong feeding} shows examples video skims and prediction results of the model. The misclassified examples demonstrate that the model does not learn the full spectrum of fine-grained information and specific features required for reliable feeding disambiguation -- the actual moment of fish-beak interaction and fish injestion may last only few frames and is a highly complex, difficult to disambiguate activity. For example, when a penguin turns quickly in a group of fish, the model may misidentify the scene. Furthermore, when a penguin rapidly dives through bubble clouds, the model may also incorrectly identify this as a predation event. These examples of misclassification indicate that non-predation events have high variability and overlap with feeding features, which makes it difficult to learn specific features for disambiguation. As stands, false classification rates are too high to significantly improve manual spotting of feeding events. Further work and larger datasets in this niche subject will be required to address this issue. However, automated processing for behaviour recognition is opening a path for detecting behaviours where more traditional approaches are  simply too slow or time-consuming. Mattern et al.~\cite{mattern}, for instance, could only process one quarter of their video data as some behaviours took too long to score and process with more traditional means.}

\vspace{-10pt}
\section{Conclusion}
\label{sec:conclusion}\vspace{-5pt}
\tb{In this paper we published and described an animal-borne underwater video dataset of penguins and introduced a ready-to-deploy deep learning system developed and benchmarked for this environment. We demonstrated that our system is capable of robustly detecting penguins (mAP50@98.0\%) in underwater scenes of the published data splits and also instances of fish (mAP50@73.3\%). We noted that the detectors benefit explicitly from learning air-bubble and fish content on both frame and object-instance level. We then extended this detector towards a dual-stream behaviour recognition network for identifying predation behaviour in penguin underwater videos. We provided detailed training settings and reflections on parameterisations and model choices. Whilst results were promising, we concluded that further work is required for useful applicability of predation behaviour detectors in field scenarios. Potential future application scenarios range widely and include prey  analysis~\cite{mattern}, social behaviour analysis for foraging~\cite{harris23}, and of course studies on energy budgets and foraging strategies~\cite{sutton2020multi}. In summary, this paper provides an annotated underwater video dataset suitable for training penguin detectors and feeding behaviour recognisers. It makes available a highly reliable underwater penguin detector, a fish detector, and a valuable first attempt towards the automated visual detection of complex behaviours in a marine predator. We publish the networks, the 'DivingWithPenguins' video dataset, annotations, splits, and weights for full reproducibility and immediate usability by practitioners. We hope that this paper and datasets can help focus the attention of the computer vision community towards researching better AI tools for animal-borne video analysis with the ultimate goal to better understand and conserve the natural world from the animal's perspective.}

\begin{table}[t]
    \centering
    \begin{tabular}{@{}lccc@{}}
    \toprule
                    & Feeding & Swimming & Average Accuracy\\ \midrule
    One layer with 512   & 76\%             & 56\%              & 66\%             \\
    Two layers with 512 & 78\%             & 78\%              & 78\%             \\ 
    Two layers with 256 & 74\%             & 92\%              & 83\%             \\ 
    Two layers with 128 & 86\%             & 64\%              & 75\%             \\ \bottomrule
    \end{tabular}\vspace{5pt}
    \caption{\tb{\textbf{Behaviour Detection Performance via Different LSTM Layouts}. \\The table shows accuracies for the two different behaviour classes under vary-\\ing depth and breadth of the semantic head.}}
    \label{tab:num_layer_hidden}
\end{table}
\begin{table} 
\tb{
    \centering \vspace{-10pt}
    \begin{tabular}{llll}
    \hline
    \multicolumn{4}{c}{\textbf{ResNet-18 Fish Frame Detector Test Accuracy per Class}}   \\ \hline
    \white{-------------------------}    &  Class Fish &  Class NoFish & Average\white{------}  \\
    Off-the-Shelf  & 90.2\%             & 91.1\%           & 90.6\%     \\
    Task-Optimised  & 91.8\%             & 90.2\%           & 91.0\%             \\ \hline
    \end{tabular}\vspace{5pt}
    \caption{ \tb{\textbf{Whole Frame Fish-Presence Classifier Performance}. Depicted are\\accuracies for an off-the-shelf and best-performing ResNet-18 network re fish\\flagging before and after optimisations. Note that despite improved overall per\\class-accuracy, improvements in fish detection trigger higher sensitivity to mis-\\classifications in NoFish frames.\vspace{-10pt}}}
    \label{tab:resnet_summary}}
\end{table}

%%%%%%%%%% end main matter %%%%%%%%%%

\vspace{-5pt}
\section{Acknowledgements and Data/Code Links}\label{sec:ack}\vspace{-5pt}
This work utilised the Blue Crystal Phase 4 supercomputer at the University of Bristol, UK. All key code, network weights, and links to the datasets can be found on GitHub at: \url{https://github.com/Kejia928/PenguinProject.git}. We thank BirdLife South Africa and all people involved in the data gathering efforts for their hard work.

%%%%%%%%%%%%%% bibliography %%%%%%%%%%
\bibliographystyle{plainnat}
\bibliography{references}

%%%%%%%%%% end bibliography %%%%%%%%%%

\end{document}